\documentclass[journal]{IEEEtran}

\usepackage{cite}
\usepackage{graphicx}
\usepackage{amsmath}
\usepackage{amssymb}
\usepackage{booktabs}
\usepackage[caption=false,font=footnotesize]{subfig}
\usepackage{xurl}
\begin{document}

\title{NBA\_Streaming: A Large-Scale Benchmark for Fine-Grained
Basketball Commentary Generation in Continuous Streams}

\author{
Lifang~Wu,~\IEEEmembership{Senior Member,~IEEE},
Yuyang~Wu,
Yangdong~Gao,
Fengyu~Liu,
Ya~Jing$^{*}$,
and~Liang~Wang,~\IEEEmembership{Fellow,~IEEE}%
\thanks{$^{*}$Corresponding author: Ya Jing.}%
\thanks{Lifang Wu, Yuyang Wu, Yangdong Gao, and Ya Jing are with the
School of Information Science and Technology, Beijing University of
Technology, Beijing 100124, China
(e-mail: \protect\url{lfwu@bjut.edu.cn};
\protect\url{yuyang_wu2003@163.com};
\protect\url{gyd2389@163.com};
\protect\url{jingya004@126.com}).}%
\thanks{Fengyu Liu is with Fudan University, Shanghai, China
(e-mail: \protect\url{liufengyu_bj@163.com}).}%
\thanks{Liang Wang is with the Institute of Automation,
Chinese Academy of Sciences, Beijing 100864, China
(e-mail: \protect\url{wangliang@nlpr.ia.ac.cn}).}%
}

\maketitle

\begin{center}
\footnotesize
This work has been submitted to the IEEE for possible publication.
Copyright may be transferred without notice, after which this version
may no longer be accessible.
\end{center}

\begin{abstract}
Live basketball commentary generation requires determining when an event
is sufficiently observable and describing it before subsequent events
unfold. However, existing methods are primarily designed for pre-segmented
clips or complete videos, making them unsuitable for continuous streams.
Existing datasets also provide limited supervision for player identities,
fine-grained actions, event attributes, and coherent event chains,
restricting the factual richness of generated commentary. To address these
limitations, we introduce NBA\_Streaming, a large-scale benchmark for
online fine-grained basketball commentary generation. It contains 307.5 hours
of basketball broadcasts and approximately 35K temporally aligned events,
with annotations of event boundaries, player identities, fine-grained
actions, event chains, and natural-language commentary. By moving from
isolated clips to continuous streams, NBA\_Streaming enables unified
evaluation of event localization, response reliability, factual grounding,
and commentary quality under causal constraints. We further propose a
causal two-stage framework that combines completion-first localization with
ball-centric semantic grounding, enabling the system to identify complete
events from observed streams and organize scene, event, identity, and action
cues for commentary generation. Extensive experiments reveal the difficulty
of NBA\_Streaming, where existing baselines struggle with online timing,
factual grounding, and fine-grained description. Our framework consistently
improves over strong alternatives, while the remaining gap highlights
NBA\_Streaming as a valuable benchmark for streaming sports video
understanding and generation. The dataset is publicly available at https://github.com/wyy-081/NBA\_Streaming.
\end{abstract}


\section{Introduction}
\label{sec:introduction}

Video captioning has advanced rapidly with recent progress in video
understanding and vision-language modeling~\cite{
yang2023vid2seq,
islam2024videorecap,
kim2024showthinktell,
wu2025eventequalized
}.
This progress has extended video-language research from generic scenes to
structured sports videos, supporting applications such as automatic
commentary generation~\cite{
yu2018finegrainedsports,
mkhallati2023soccernetcaption,
qi2023goal,
rao2024matchtime
}.
Compared with generic captioning, sports commentary requires a deeper
understanding of domain-specific semantics, including player identities,
fine-grained actions, outcomes, and surrounding game context. Automatically
generating such commentary is valuable for intelligent broadcasting,
accessible sports viewing, and large-scale game analysis.

Despite recent progress, existing methods remain limited in realistic
streaming basketball scenarios. First, most approaches~\cite{
xi2025player,
xi2024knowledge
}
are designed for offline settings, where the full video is available for
analysis, making them unsuitable for real-time applications that require
online response. Second, existing datasets~\cite{
xu2016msrvtt,
zhang2024bhcommentary,
yu2018finegrainedsports
}
mainly focus on clip-level video understanding or captioning tasks. They do
not simultaneously provide continuous game streams and event-level temporal
boundaries. Beyond these limitations, current methods also suffer from
insufficient semantic granularity. Existing basketball video
descriptions~\cite{
wu2022nsva,
zhang2024bhcommentary,
xi2025player
}
are typically limited to coarse event categories and lack fine-grained
semantic elements such as detailed actions, player identities, and event
chains. As a result, they cannot meet the requirements of professional
basketball commentary, which requires not only identifying what happens but
also generating commentary with coherent logical relations.

\begin{figure*}[!t]
    \centering
    \includegraphics[width=\textwidth]{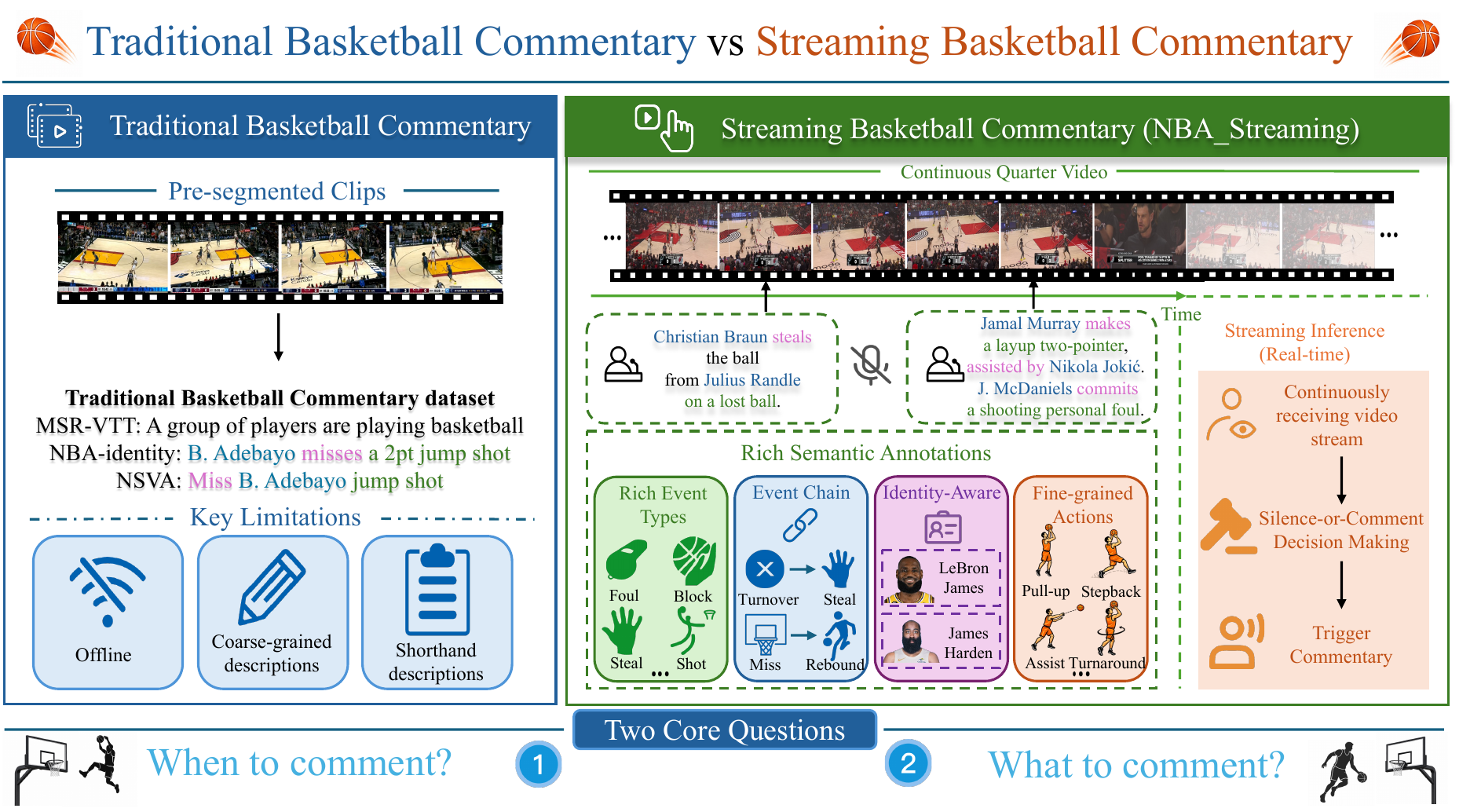}
   \caption{Comparison between traditional clip-based basketball captioning
and streaming basketball commentary. Unlike pre-segmented offline settings,
NBA\_Streaming operates on continuous quarter-long videos with rich event,
identity, and fine-grained action annotations, requiring the model to
jointly determine when to comment and what to comment in real time.}
    \label{fig:task}
\end{figure*}

To address this gap, we introduce NBA\_Streaming, a large-scale benchmark
for fine-grained streaming basketball commentary generation. As illustrated
in Fig.~\ref{fig:task}, the task continuously processes an incoming game
video and addresses two questions: \emph{when to comment} and
\emph{what to comment}. The benchmark provides full-game video streams,
event boundaries, player identities, action details, and contextual event
relations, enabling evaluation of event localization, response timeliness,
and commentary quality under causal constraints.

Based on this benchmark, we propose a causal two-stage framework to address
two challenges: online event localization and commentary generation with
rich event details. Stage I adopts completion-first localization: it first
determines whether an event has just ended and then selects the most complete
event segment from the observed history. Since ball motion naturally
connects event types, involved players, and fine-grained actions in
basketball, Stage II performs ball-centric semantic grounding to organize
these cues and generate more accurate commentary.

The main contributions of this paper are summarized as follows:

\begin{itemize}
    \item We formulate the task of Fine-Grained Basketball Commentary
    Generation in Continuous Streams and introduce NBA\_Streaming. This
    benchmark enables evaluation of two core model capabilities: online
    event localization and basketball commentary generation with rich event
    details.

    \item Our causal two-stage framework introduces two innovations:
Stage I uses event-completeness supervision to accurately localize
complete event segments. Stage II uses ball-centric semantic grounding
to progressively model event classification, player identities, and
fine-grained actions. Notably, player identity recognition is learned in
a weakly supervised manner without introducing additional identity
annotations or external knowledge.

    \item Experiments show clear gains in temporal localization and
    commentary quality, while NBA\_Streaming provides a challenging testbed
    for Fine-Grained Basketball Commentary Generation in Continuous Streams.
\end{itemize}
\section{Related Work}
\label{sec:related_work}

\subsection{Sports Video Captioning}

Sports video captioning~\cite{
kim2024showthinktell,
xi2025simple
}
aims to generate domain-specific textual descriptions for sports videos.
Compared with generic video captioning, it requires models to capture not
only visual content, but also action processes, event outcomes, and
structured sports semantics. Related studies have explored hierarchical and concept-aware video captioning~\cite{gao2021hierarchical},~\cite{yang2023concept}, as well as long-term athlete association in sports videos~\cite{kong2020long}.

In basketball, EAC~\cite{xi2024knowledge},
LLM-IAVC~\cite{xi2025player}, and
BaskIDCap~\cite{li2026novel}
leverage external knowledge or separately annotated player regions to
generate identity-aware descriptions. However, these designs incur
substantial annotation costs.

Existing basketball commentary benchmarks, including
VC-NBA-2022~\cite{xi2025simple},
BH-Commentary~\cite{zhang2024bhcommentary},
NBA-Identity~\cite{xi2025player}, and
NSVA~\cite{wu2022nsva},
are also primarily clip-based. In contrast, our work directly processes
continuous game streams and jointly models event localization, player
identities, fine-grained actions, and natural-language commentary
generation, while learning player identities under weak supervision
without requiring annotated player regions.

\subsection{Online Temporal Action Localization}

Online temporal action localization detects action instances from
streaming videos without accessing future, a causal setting also considered in online action segmentation~\cite{jiang2026egocentric}. FineAction~\cite{liu2022fineaction} highlights the challenges of localizing fine-grained and densely occurring actions with ambiguous temporal boundaries. Existing methods,
including MATR~\cite{song2024online},
HAT~\cite{reza2024hat}, and
OnPoint~\cite{reza2026onpoint},
satisfy the causal observation constraint, but often involve boundary
regression and online suppression or merging. Consequently, the prediction
submission time may not align with the actual event completion time.

In contrast, our method immediately commits an event upon the first valid
end trigger, without delayed merging or subsequent boundary revision.

\subsection{Streaming Video Understanding}

{\sloppy
Existing streaming video understanding methods, including
Live Video Captioning~\cite{blancofernandez2025livevideocaptioning},
Streaming Dense Video Captioning~\cite{zhou2024streamingdensecaptioning},
LiveCC~\cite{chen2025livecc},
OVBench~\cite{huang2025online},
VideoLLM-Online~\cite{chen2024videollmonline},
long-term memory approaches~\cite{chatterjee2025streaming,zhang2025flash},
and StreamMind~\cite{ding2025streammind}
aim to perform online perception, memory updating, and real-time response
over continuously arriving video streams without accessing future frames.
}

However, these works mainly focus on open-domain video understanding,
online dialogue, and real-time response. Fine-Grained Basketball
Commentary Generation in Continuous Streams introduces additional
challenges: it requires not only timely event triggering and accurate
temporal localization, but also domain-specific commentary grounded in
fine-grained basketball semantics.

\section{Dataset}
\label{sec:dataset}

\subsection{Data Construction}
\label{subsec:data_construction}

Play-by-play (PBP) records are chronological logs that describe game
events and their corresponding game-clock timestamps. NBA Streaming is constructed from official NBA PBP records and full broadcast videos of 152 games from the 2025–26 NBA season. Since PBP records use the game clock
while broadcast videos follow an independent timeline, we apply OCR to
recognize the on-screen game clock and match it with the official
timestamps. NBA\_Streaming preserves continuous full-game videos,
supporting online event perception and commentary generation.

\begin{figure}[!t]
    \centering
    \includegraphics[width=\columnwidth]{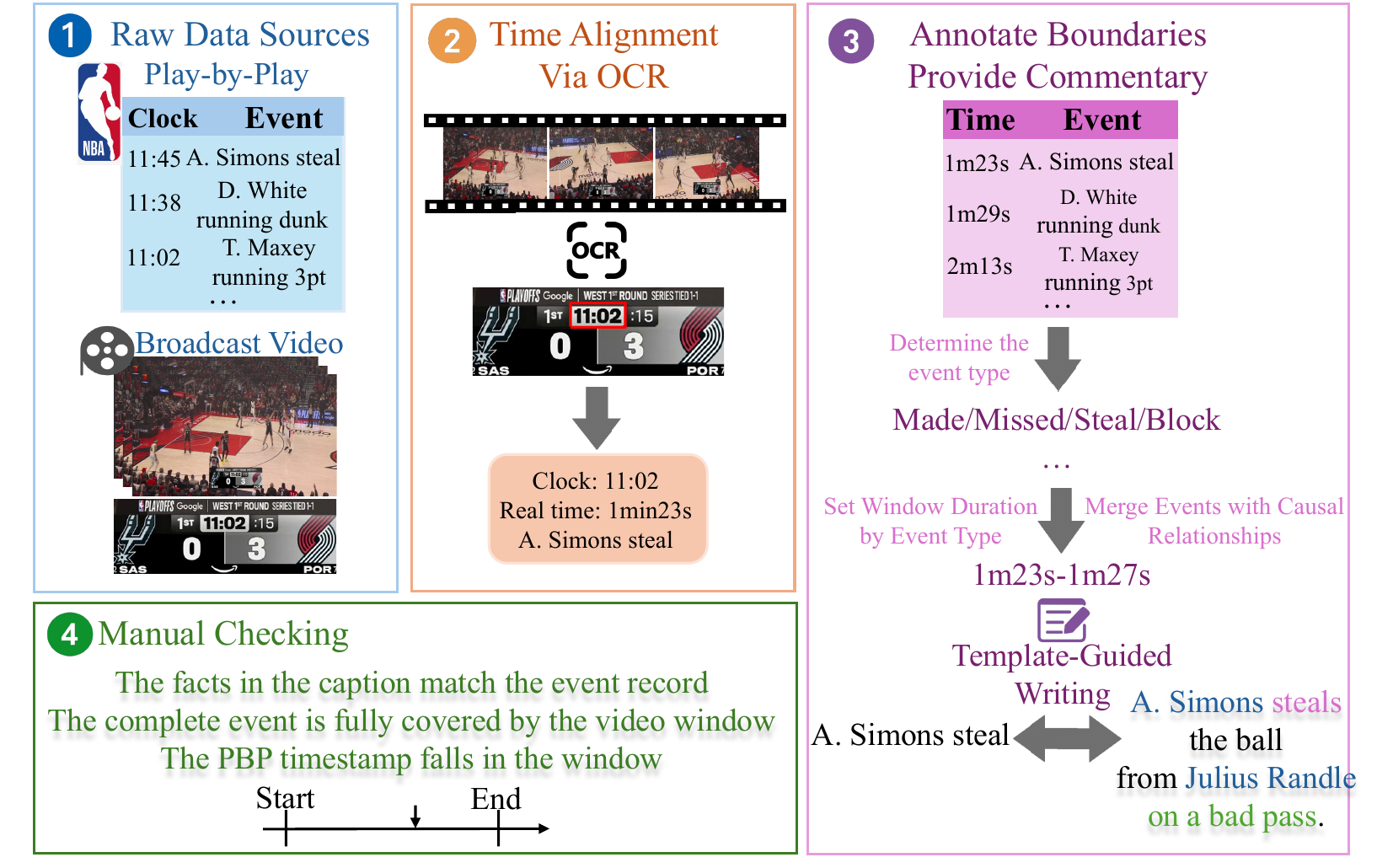}
    \caption{Construction pipeline of NBA\_Streaming. The pipeline integrates
    play-by-play records and broadcast videos, aligns event timestamps with
    video time via OCR, annotates event boundaries and commentary, and merges
    causally related adjacent events into coherent event chains. Manual
    checking is then performed to ensure temporal coverage and semantic
    consistency.}
    \label{fig:dataset_construction}
\end{figure}

As shown in Fig.~\ref{fig:dataset_construction}, three annotators with
basketball expertise reviewed representative games and jointly
established event-type-specific boundary rules. For each discrete PBP
timestamp, they set the start and end boundaries according to the event
type, thereby expanding the timestamp into a complete event window. We
further merge the boundaries of related events.

Since the original PBP descriptions consist of brief abbreviations and
cannot be used directly as natural-language commentary, we use
Llama3-8B~\cite{grattafiori2024llama} to expand the PBP descriptions into natural commentary
according to templates designed for different event types.

To quantify annotation quality, we randomly sample 20 games and evaluate
three aspects: whether the PBP timestamp is included within the annotated
interval, whether the expanded commentary is semantically consistent
with the original PBP description and whether the annotated boundaries
cover the complete event process. The pass rates for timestamp
inclusion, commentary consistency, and complete-event coverage are
97\%, 99\%, and 96\%, respectively.

\subsection{Dataset Statistics}
\label{subsec:dataset_statistics}

NBA\_Streaming contains 152 full-game replays totaling approximately
307.5 hours, with 35K events and 50.4K commentary sentences. To support
systematic training and evaluation, we perform a game-level partition,
using 121 games for training, 15 for validation, and 16 for testing.

\begin{figure*}[!t]
    \centering
    \includegraphics[width=\textwidth]{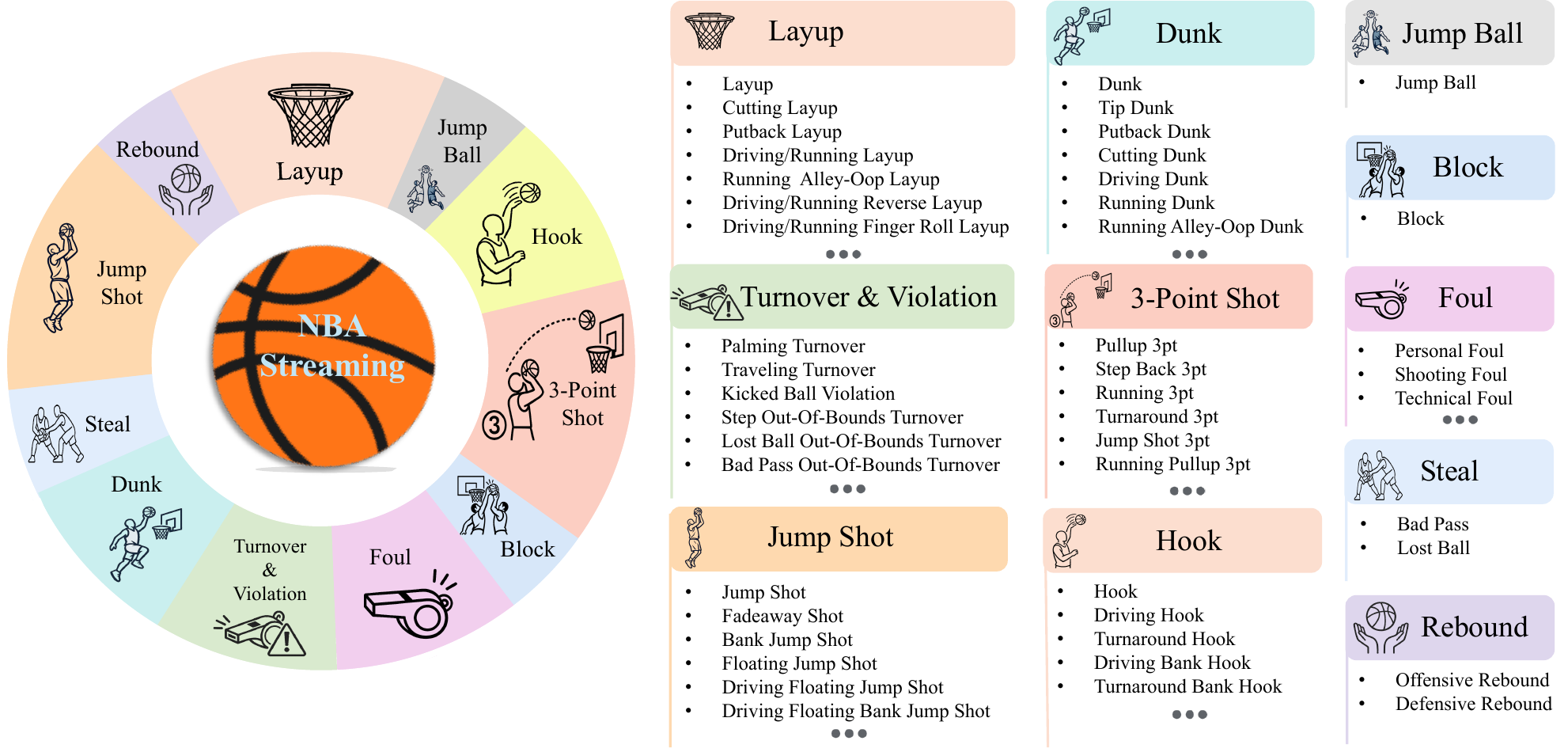}
    \caption{Fine-grained action taxonomy of NBA\_Streaming. The taxonomy
    organizes basketball actions into 11 major action families and further
    divides them into 71 fine-grained action categories, covering diverse
    shooting, rebounding, defensive, foul, turnover, and other basketball
    actions.}
    \label{fig:action_taxonomy}
\end{figure*}


As shown in Fig.~\ref{fig:action_taxonomy}, the dataset covers 11 major
action families, including layups, jump shots, three-point shots, dunks,
hook shots, rebounds, steals, fouls, blocks, turnovers/violations, and
jump balls. These major families are further divided into a total of
71 fine-grained action categories, providing detailed supervision for
semantic understanding and commentary generation.

At the event level, NBA\_Streaming further defines 15 event types:
2pt\_\allowbreak made, 2pt\_\allowbreak missed,
2pt\_\allowbreak missed\_\allowbreak rebound,
3pt\_\allowbreak made, 3pt\_\allowbreak missed,
3pt\_\allowbreak missed\_\allowbreak rebound,
putback\_\allowbreak chain, steal, turnover, rebound, foul,
shot\_\allowbreak foul\_\allowbreak chain,
block\_\allowbreak chain, jump\_\allowbreak ball, and violation.
These labels characterize the overall semantic category of each event.

The identity annotations cover all 30 NBA teams and contain 539 distinct
player identities, providing comprehensive identity supervision for
player-aware event understanding and commentary generation.

\begin{table*}[!t]
    \centering
    \caption{Comparison of NBA\_Streaming with existing video-language
    datasets.}
    \label{tab:dataset_comparison}
    \resizebox{\textwidth}{!}{
    \begin{tabular}{lccccccc}
        \toprule
        Dataset
        & \#Events
        & \#Sent.
        & Dur. (h)
        & Avg. Words
        & Streaming
        & Identity
        & Semantic Granularity \\
        \midrule

        MSR-VTT~\cite{xu2016msrvtt}
        & 10.0k
        & 200.0k
        & 41.2
        & 9.2
        & $\times$
        & $\times$
        & 1 \\

        FSN~\cite{yu2018finegrainedsports}
        & 2.0k
        & 6.5k
        & --
        & --
        & $\times$
        & $\times$
        & 2 \\

        NSVA~\cite{wu2022nsva}
        & 32.0k
        & 44.6k
        & 84.8
        & 6.5
        & $\times$
        & $\checkmark$
        & 4 \\

        BH-Commentary~\cite{zhang2024bhcommentary}
        & 4.3k
        & 4.3k
        & 10.1
        & --
        & $\times$
        & $\times$
        & 3 \\

        VC-NBA-2022~\cite{xi2025simple}
        & 3.9k
        & 3.9k
        & --
        & --
        & $\times$
        & $\checkmark$
        & 3 \\

        NBA-Identity~\cite{xi2025player}
        & 9.7k
        & 9.7k
        & 8.9
        & --
        & $\times$
        & $\checkmark$
        & 3 \\

        \textbf{NBA\_Streaming}
        & \textbf{35.0k}
        & \textbf{50.4k}
        & \textbf{307.5}
        & \textbf{13.2}
        & $\checkmark$
        & $\checkmark$
        & \textbf{5} \\

        \bottomrule
    \end{tabular}}
\end{table*}

As shown in Table~\ref{tab:dataset_comparison}, NBA\_Streaming surpasses
existing basketball datasets in both the number of events and total
video duration. It also provides 50.4K commentary sentences with an
average length of 13.2 words, reflecting richer natural-language
descriptions.

We also introduce a taxonomy of semantic granularity to systematically
compare the semantic levels of captions across different datasets.
Specifically, captions are evaluated along five dimensions, each assigned
a binary score of 0 or 1:

\begin{itemize}
    \item \textbf{Specific basketball action:}
    whether the caption explicitly describes a basketball action or
    event type, rather than providing only a generic scene-level
    description;

   \item \textbf{Player name:}
whether the caption explicitly includes specific player names rather than
using generic references such as ``a player'';

    \item \textbf{Complex event details:}
    whether the caption captures detailed event types, such as
    turnovers, violations, fouls, and their underlying causes;

    \item \textbf{Complete event chain:}
    whether the caption describes multiple temporally related events,
    forming a coherent event sequence, such as
    shot $\rightarrow$ miss $\rightarrow$ rebound;

    \item \textbf{Fluent natural language:}
    whether the caption is expressed as a fluent natural-language
    sentence, rather than as shorthand descriptions.
\end{itemize}

\subsection{Ethics Statement}
\label{subsec:ethics_statement}

The NBA\_Streaming dataset is released under the Creative Commons
Attribution-NonCommercial 4.0 International License (CC BY-NC 4.0),
restricting its use to academic and research purposes while permitting
free use, modification, and distribution with proper attribution and
strictly prohibiting any commercial use.

\section{Method}
\label{sec:method}

\subsection{Task Definition}
\label{subsec:task_definition}

At decision step $t$, the model observes only the video prefix
$X_{\leq t}=\{x_1,\ldots,x_t\}$, where $x_i$ is the $i$-th frame
and future frames are unavailable. The commentary decision is
\begin{equation}
y_t=
\mathbb{I}
\left[
\mathcal{G}_{e}(X_{\leq t})>\tau
\right],
\label{eq:event_trigger}
\end{equation}
where $\mathcal{G}_{e}$ is the event-ending detector,
$\mathcal{G}_{e}(X_{\leq t})$ is the predicted ending probability,
$\tau$ is the trigger threshold, and $\mathbb{I}[\cdot]$ is the
indicator function. Here, $y_t=0$ denotes silence and $y_t=1$
denotes an event trigger.

At the first trigger, the current step is fixed as the event end, the
start is recovered from the observed history, and commentary is generated
from the resulting event:
\begin{equation}
\begin{gathered}
\hat{t}_e = t, \\
\hat{t}_s = \mathcal{G}_{s}(X_{\leq t},\hat{t}_e), \\
\hat{c} =
\mathcal{H}\left(
X_{\hat{t}_s:\hat{t}_e},
\mathcal{P}(\hat{e})
\right),
\end{gathered}
\label{eq:event_commentary_generation}
\end{equation}
where $\hat{t}_e$ and $\hat{t}_s$ are the predicted event end and
start, respectively. $\mathcal{G}_{s}$ is the completeness-aware causal
start localizer, which selects the most complete event interval from
end-anchored historical candidates. The resulting event is denoted by
$\hat{e}=(\hat{t}_s,\hat{t}_e)$, and
$X_{\hat{t}_s:\hat{t}_e}$ is its corresponding video segment.
$\mathcal{P}(\hat{e})$ denotes the visual and semantic prompts extracted
from the event, $\mathcal{H}$ is the commentary generator, and
$\hat{c}$ is the generated commentary.

\subsection{Architecture Design}
\label{subsec:architecture_design}

\begin{figure*}[!t]
    \centering
    \includegraphics[width=\textwidth]{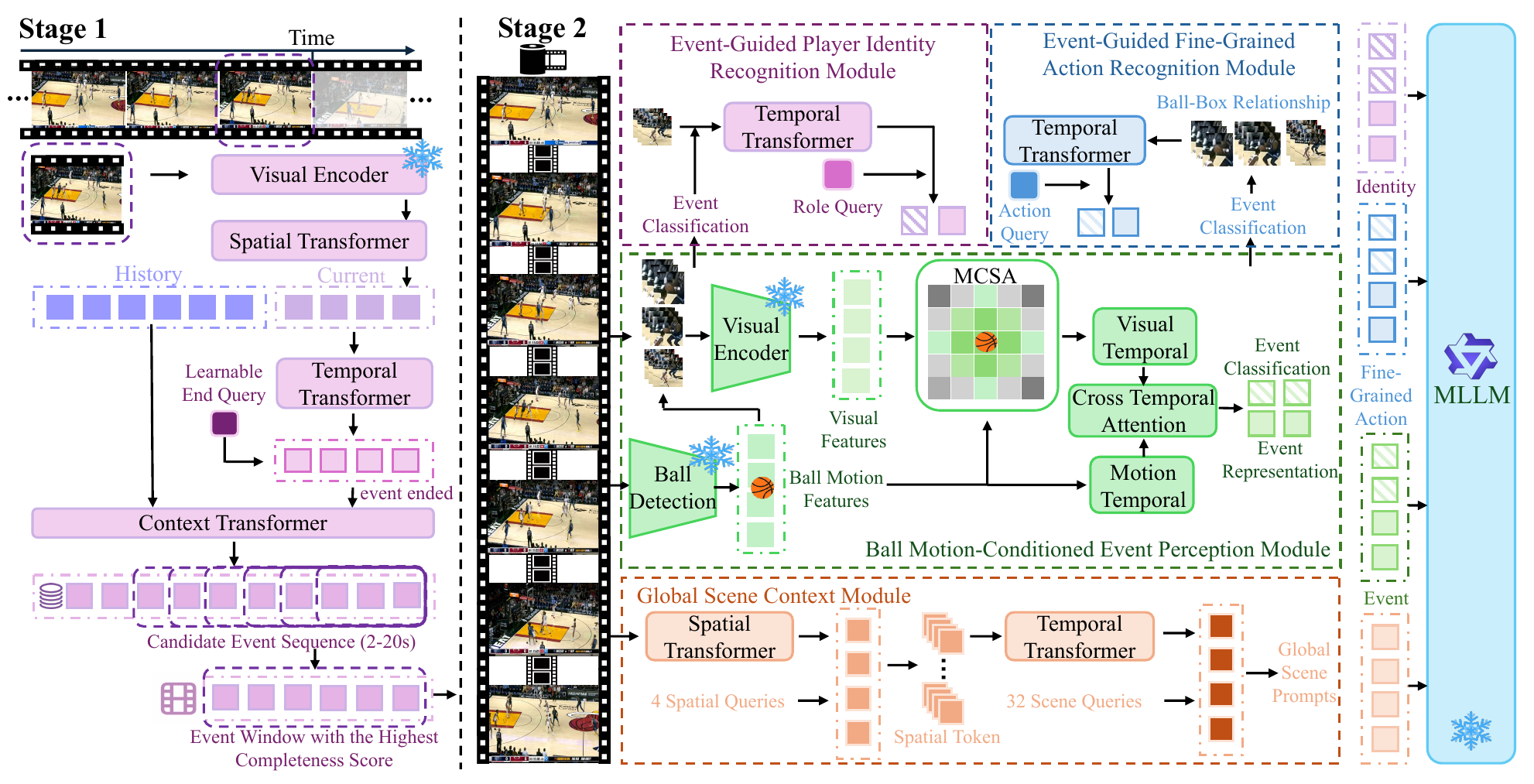}
   \caption{Overview of the proposed causal two-stage framework for streaming
basketball commentary generation. Stage I performs completion-first
localization to identify complete events from the observed video stream.
Stage II performs ball-centric semantic grounding to progressively organize
scene, event, player identity, and fine-grained action cues for accurate
commentary generation.}
    \label{fig:framework}
\end{figure*}

Fig.~\ref{fig:framework} shows our causal two-stage framework for
complete-event localization and ball-centric semantic grounding.

\subsubsection{Stage I: Completeness-Aware Causal Event Localization}

Rather than predicting the start and end boundaries independently,
Stage I identifies a complete event interval from the observed stream.
This preserves the complete event chain. The spatial tokens from the visual
encoder are aggregated through learnable spatial attention and projected
into frame-level features. A temporal Transformer encodes the current
observation window, using a learnable end query to determine whether the
event has ended. When the end probability first exceeds the threshold,
the model immediately fixes the current timestamp as the event endpoint.
The historical memory and current observation are then encoded by a context
Transformer, and multiple candidates of different durations ending at this
timestamp are constructed. Each candidate integrates its internal interval
content, boundary context, and duration representation. The completeness
head evaluates the event completeness of each candidate. During inference,
the candidate interval with the highest completeness score is selected as
the event interval.

\subsubsection{Stage II: Conditional Semantic Pyramid}

Since event types, player identities, and fine-grained actions directly
affect the quality of basketball commentary, we design four complementary
modules to model these semantics. M0 provides global scene context. M1
focuses on ball motion to recognize event types and event chains, serving
as the semantic foundation for M2 player identity recognition and M3
fine-grained action recognition.

\paragraph{M0: Global Scene Context Module}

A one-layer spatial Transformer first models relationships among the
spatial tokens extracted by the visual encoder. Four learnable frame
queries then cross-attend to the encoded spatial tokens, producing four
compact representations for each frame. These representations are
concatenated over time and processed by a three-layer temporal Transformer
to capture event evolution. Finally, 32 learnable scene queries retrieve
global information from the temporal sequence through two query layers,
each containing query self-attention and cross-attention. The resulting
scene tokens are projected into the MLLM's language-embedding space.

\paragraph{M1: Ball Motion-Conditioned Event Perception Module}

M1 recognizes event semantics from ball motion and nearby visual features.
At frame \(t\), the frozen WASB detector~\cite{tarashima2023widely}
provides the ball position \(\mathbf{p}_t\) and confidence \(c_t\). We
derive the validity indicator \(g_t\) and construct

\begin{equation}
\mathbf{m}_t=
[\mathbf{p}_t,\mathbf{v}_t,\|\mathbf{v}_t\|,
\mathbf{a}_t,\|\mathbf{a}_t\|,c_t,g_t],
\label{eq:ball_motion_feature}
\end{equation}

where \(\mathbf{v}_t\) and \(\mathbf{a}_t\) are the ball velocity and
acceleration.
Centered on the ball position, the model extracts visual information at
  three progressively larger scales: a local feature describing the ball
  itself, a contextual feature capturing its interaction with the
  surroundings, and a large-scale feature covering the potential ball
  handler. Motion-Conditioned Spatial Attention uses the ball-motion
  representation as the query to dynamically fuse these three visual
  features according to the current ball motion, producing a frame-level
  event visual representation.

  The ball-motion sequence and event-visual sequence are then encoded by
  separate temporal Transformers. Cross-temporal attention establishes the
  correspondence between ball motion and visual changes. Finally, a
  learnable event query aggregates the fused sequence into an event
  representation. The predicted event category and event representation
  serve as the textual and visual event prompts, respectively, and also
  guide M2 and M3.

\paragraph{M2: Event-Guided Player Identity Recognition Module}

M2 uses the event type predicted by M1 to determine the player role to be
  recognized and the temporal interval relevant to that role. For example,
  shooter identification focuses on observations before the ball begins to
  rise. At each selected timestamp, M2 reuses the largest-scale feature from
  M1, which covers the potential ball handler, without invoking an
  additional player detector or tracker.

At selected timestamps, M2 extracts the player appearance from the
large-scale region to identify the player. Meanwhile, ball motion and the
spatial relationship between the ball and this region are used to determine
which timestamps are most relevant to the target role. A temporal
Transformer encodes these observations into a candidate sequence. The
model then learns a query for each player role to assign attention weights
across the candidate timestamps. The player appearance features are
aggregated according to these weights, producing the identity
representation of the target role.

  Since the commentary provides player names and their roles but does not
  specify the corresponding video timestamps, the role-related observation
  sequence is treated as a weakly supervised set from which the target
  identity must be inferred. Finally, the predicted role and player name
  form the textual identity prompt, while the aggregated identity
  representation is projected into a continuous identity prompt.

\paragraph{M3: Event-Guided Fine-Grained Action Recognition}

M3 uses the event type predicted by M1 to determine the fine-grained action category and its relevant temporal phase. It directly reuses the ball-motion features and the three ball-centered visual features extracted by M1, including local ball appearance, surrounding interaction context, and a
  large-scale region covering the ball handler. M3 further models the player's visual changes within the large-scale region and its spatial
  relationship with the ball using a temporal Transformer. Each action category is associated with a learnable action query, which attends to the
  corresponding key phase and aggregates its features into an action representation. Finally, the recognized category and aggregated representation
  are respectively used as the textual and visual action prompts for the MLLM.

\subsection{Optimization Objectives}
\label{subsec:optimization_objectives}

Stage I uses Focal Loss~\cite{lin2017focal} for event-ending detection and
binary cross-entropy for candidate completeness, with temporal IoU serving
as the completeness target for each candidate. M1 uses weighted cross-entropy
for the 15 event categories and five auxiliary event attributes: shot or
non-shot, two or three-point attempt, made or missed result, post-shot state,
and non-shot event type. M2 combines a weak set-level identity loss over all
ball-anchored observations, a second identity classification loss to the
identity representation selected by the corresponding role query, and a
supervised contrastive loss~\cite{khosla2020supervised} that pulls together
appearance representations of the same player while separating those of
different players. M3 combines standard and class-balanced
cross-entropy~\cite{cui2019class} for both shot attributes and fine-grained
action categories.
\section{Experiments}
\label{sec:experiments}

\subsection{Experimental Settings}
\label{subsec:experimental_settings}

\subsubsection{Dataset and Baselines}

All experiments are conducted on NBA\_Streaming using the same data split,
6-FPS input, and three NVIDIA A100 GPUs. We adapt and retrain all baselines
on our dataset, except for the standalone frozen Qwen3-VL-8B-Instruct
model~\cite{bai2025qwen3}. StreamMind~\cite{ding2025streammind} and
VideoLLM-Online~\cite{chen2024videollmonline} are adopted as streaming
video language baselines. MATR~\cite{song2024online},
HAT~\cite{reza2024hat}, and OnPoint~\cite{reza2026onpoint} are adopted as
Stage-I online localization baselines. For Stage II,
Qwen3-VL-8B-Instruct~\cite{bai2025qwen3} is used as the commentary
generator. In these two-stage settings, we train a dedicated visual
projector and soft prompts, and convert the event type predicted in Stage I
into a textual prompt.

We also separately evaluate the native capability of frozen
Qwen3-VL-8B-Instruct~\cite{bai2025qwen3} using only a manually designed
instruction prompt. In this zero-shot setting, the model directly relies
on its original visual encoder and pretrained multimodal reasoning and
generation capabilities, without any additional training or task-specific
adaptation on NBA\_Streaming.

MatchTime~\cite{rao2024matchtime}, originally designed for soccer
commentary generation, and IAVC~\cite{xi2025player}, a basketball
commentary method with explicit player-identity awareness, are further
adapted as sports commentary baselines using MATR-localized segments.
These baselines are evaluated in terms of response timing, boundary
localization, semantic recognition, and overall commentary quality.

\subsubsection{Implementation Details}

We employ a frozen CLIP~\cite{radford2021learning} encoder to extract
visual features. In Stage I, the current observation window and historical
memory contain 60 and 72 frames, respectively, providing 22 seconds of
context. The event-ending threshold is set to \(0.5\). We construct 37
end-anchored candidates with durations from \(2\) to \(20\) seconds at
\(0.5\)-second intervals. Stage I is trained for 50
epochs using AdamW~\cite{loshchilov2017decoupled}, with a global batch size
of 576 and an initial learning rate of \(1\times10^{-4}\). In Stage II,
each localized event retains at most 128 frames. M1, M2, and M3 are trained
for 15, 30, and 20 epochs with batch sizes of 512, 768, and 512,
respectively. All modules are optimized using
AdamW~\cite{loshchilov2017decoupled}. During final commentary training,
CLIP~\cite{radford2021learning},
Qwen3-VL-8B-Instruct~\cite{bai2025qwen3}, and the pretrained M1--M3 modules
are frozen, while M0 and visual projectors are optimized. The model is
trained for up to 12 epochs with a batch size of 4 per GPU.

\subsection{Evaluation Metrics}
\label{subsec:evaluation_metrics}

We evaluate the model in terms of temporal localization, commentary
generation, and runtime efficiency. For Stage II, we evaluate only events
whose start and end boundary errors are both within 2 seconds, with further
comparisons on the common event subset.

\subsubsection{Temporal Localization Ability}

We report Start $\Delta t$ and End $\Delta t$, which measure
the mean absolute errors between the predicted
and ground-truth start and end boundaries,
respectively. For matching, a predicted event is considered
correctly matched to a ground-truth event only
when both its start and end boundaries fall
within $\pm 2$ s of the corresponding ground-truth
boundaries, and all such matches are
established in a one-to-one manner. Event Response corresponds
to recall~\cite{sokolova2009systematic}, measuring the proportion of annotated
events that receive a correct response, whereas Correct Response
corresponds to precision~\cite{sokolova2009systematic}, measuring the
proportion of system responses correctly matched to annotated events.

\subsubsection{Commentary Generation Quality}

Linguistic quality is evaluated using CIDEr~\cite{vedantam2015cider} and
BLEU-4~\cite{papineni2002bleu}, with both scores multiplied by 100. Factual
correctness is measured by Identity Accuracy, Event Type Accuracy, and
Fine-grained Action Accuracy, which evaluate player names, 15 event types,
and individual action terms in the commentary, respectively.

\subsubsection{Runtime Efficiency}

To evaluate real-time efficiency, we measure throughput on the
  full test set under a unified three-GPU protocol. Each GPU processes one
  independent video stream, and no stream is split across GPUs. For each
  method, the measured pipeline covers raw-video decoding, its native feature
  extraction, online event localization, and commentary generation. Aggregate
  throughput is computed as the total number of frames processed by all three
  GPUs divided by the wall-clock time of the slowest GPU. Since the three
  streams are each supplied at 6 FPS, their combined input rate is 18 FPS;
  therefore, an aggregate throughput above 18 FPS indicates real-time
  processing of all three streams.

\begin{figure*}[!t]
    \centering
    \includegraphics[width=\textwidth]{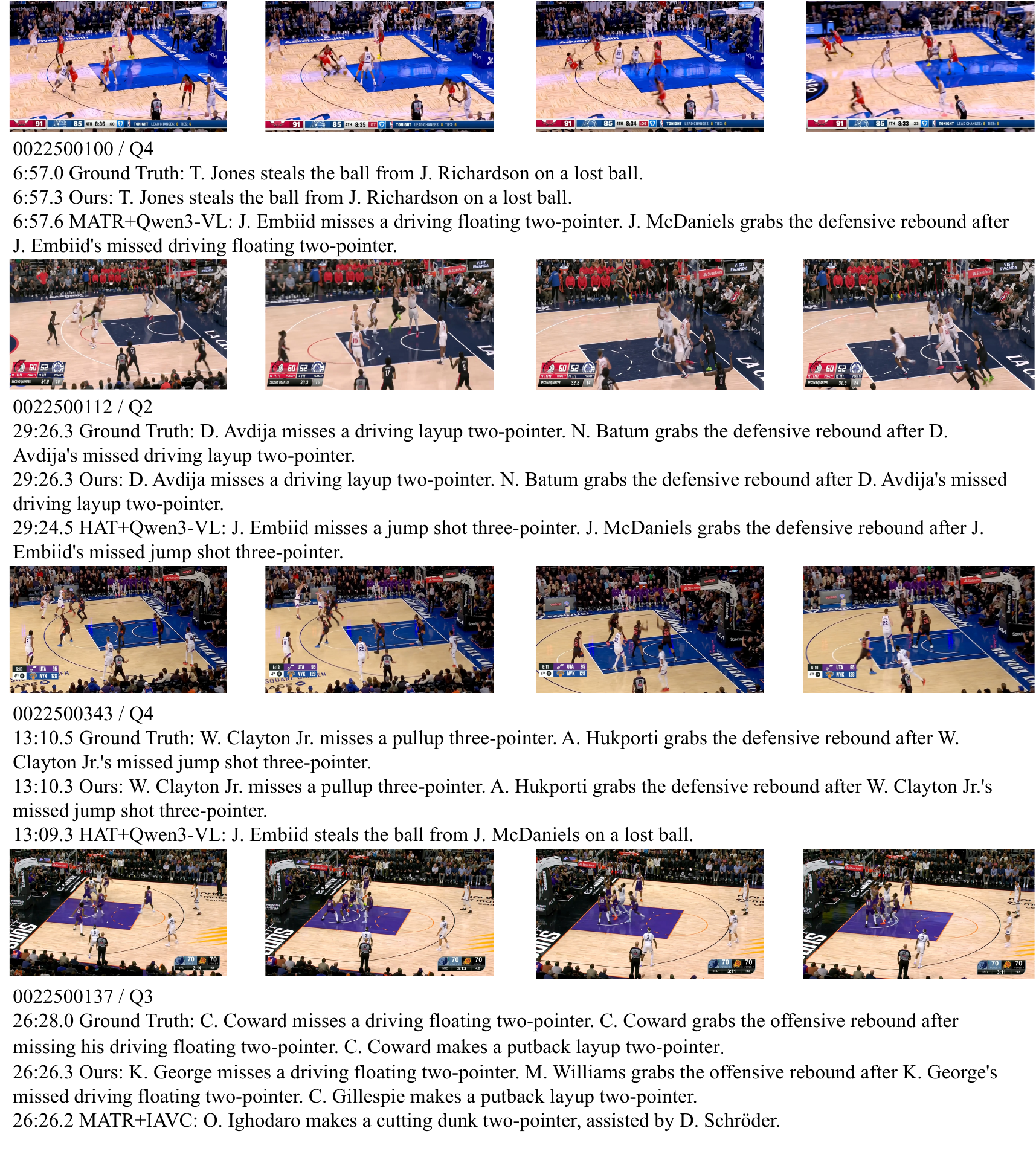}
    \caption{Qualitative comparison on four representative basketball events,
    highlighting event understanding, player identity recognition, and
    fine-grained action description under different event complexities.}
    \label{fig:qualitative_results}
\end{figure*}

\subsection{Results on NBA\_Streaming}
\label{subsec:results}

\begin{table*}[!t]
    \centering
    \caption{Comparison of online event localization, commentary
    generation, and efficiency with existing methods. C denotes CIDEr,
    while Id., Evt., and Act. denote identity, event-type, and action
    accuracies (\%), respectively.}
    \label{tab:main_results}
    \scriptsize
    \renewcommand{\arraystretch}{1.15}
    \setlength{\tabcolsep}{2.0pt}
    \resizebox{\textwidth}{!}{%
    \begin{tabular}{lcccccccccc}
        \toprule
        Method
        & Start $\Delta t$ (s) $\downarrow$
        & End $\Delta t$ (s) $\downarrow$
        & Event Response $\uparrow$
        & Correct Response $\uparrow$
        & C $\uparrow$
        & B-4 $\uparrow$
        & Id. $\uparrow$
        & Evt. $\uparrow$
        & Act. $\uparrow$
        & FPS $\uparrow$ \\
        \midrule

        Qwen3-VL~\cite{bai2025qwen3}
        & 56.360 & 56.120 & 0.0141 & 0.0236
        & 16.091 & 4.503 & 2.532 & 55.769 & 3.279 & 11.73 \\

        VideoLLM-Online~\cite{chen2024videollmonline}
        & -- & 5.663 & 0.5329 & 0.4443
        & 12.236 & 7.965 & 0.110 & 8.316 & 15.071 & 52.39 \\

        MATR~\cite{song2024online} + Qwen3-VL~\cite{bai2025qwen3}
        & 5.474 & 5.057 & 0.4915 & 0.3618
        & 42.864 & 17.651 & 0.088 & 46.116 & 22.397 & 93.64 \\

        HAT~\cite{reza2024hat} + Qwen3-VL~\cite{bai2025qwen3} 
        & 16.682 & 16.509 & 0.3650 & 0.3728
        & 37.257 & 20.569 & 0.000 & 44.585 & 29.447 & 95.86 \\

        MATR~\cite{song2024online} + MatchTime~\cite{rao2024matchtime} 
        & 5.474 & 5.057 & 0.4915 & 0.3618
        & 73.772 & 19.110 & 4.542 & 46.006 & 25.732 & 78.96 \\

        StreamMind~\cite{ding2025streammind} 
        & -- & 31.867 & 0.4549 & 0.2722
        & 14.902 & 4.423 & 0.375 & 17.586 & 4.933 & 158.51 \\

        MATR~\cite{song2024online} + IAVC~\cite{xi2025player} 
        & 5.474 & 5.057 & 0.4915 & 0.3618
        & 50.981 & 17.741 & 2.067 & 45.590 & 23.914 & 75.74 \\

        OnPoint~\cite{reza2026onpoint} + Qwen3-VL~\cite{bai2025qwen3} 
        & 3.998 & 3.359 & 0.2613 & 0.0915
        & 27.287 & 11.263 & 0.457 & 27.254 & 13.018 & 83.07 \\

        \textbf{Ours}
        & \textbf{2.470}
        & \textbf{2.094}
        & \textbf{0.6539}
        & \textbf{0.5910}
        & \textbf{248.355}
        & \textbf{41.166}
        & \textbf{28.533}
        & \textbf{71.387}
        & \textbf{47.053}
        & \textbf{230.65} \\

        \bottomrule
    \end{tabular}%
    }
\end{table*}

As shown in Table~\ref{tab:main_results}, our method achieves the lowest
boundary errors, with Start $\Delta t$ and End $\Delta t$ of 2.470\,s and
2.094\,s, reducing the corresponding errors of OnPoint by 38.2\% and
37.7\%, respectively. It also obtains the highest Event Response and
Correct Response scores of 0.6539 and 0.5910, indicating broader coverage
of ground-truth events and fewer incorrect responses. These improvements
mainly result from the explicit completeness supervision in Stage I, which
enables more accurate event localization. Among matched events whose start
and end boundary errors are both within 2\,s, our method outperforms the
best-performing baseline on each metric by 174.583 CIDEr points, 20.597
BLEU-4 points, 23.991 percentage points in Identity Accuracy, and 17.606
percentage points in Fine-grained Action Accuracy. It also improves Event
Accuracy by 25.271 percentage points over the strongest two-stage baseline.
These results demonstrate the effectiveness of ball-centric semantic
grounding. Although Qwen3-VL~\cite{bai2025qwen3} achieves seemingly high
Event Accuracy, it matches only 52 events and its predictions largely
collapse to 2pt-miss, making the result unrepresentative. Finally, our
complete model achieves a throughput 230.65 FPS, exceeding the fastest
baseline by 72.14 FPS. This efficiency mainly benefits from our
event-driven two-stage design and feature reuse, whereas Qwen3-VL directly
uses its native visual encoder and full multimodal inference pipeline, and
the other baselines retain their respective feature-extraction and
inference pipelines.

\subsubsection{Comparison on Commonly Matched Events}

\begin{table}[!t]
    \centering
    \caption{Comparison on the events commonly matched by all methods.
    C denotes CIDEr, while Id., Evt., and Act. denote identity,
    event-type, and fine-grained action accuracies (\%), respectively.
    Higher values indicate better performance for all metrics.}
    \label{tab:common_events}
    \scriptsize
    \renewcommand{\arraystretch}{1.10}
    \setlength{\tabcolsep}{2.5pt}
    \resizebox{\columnwidth}{!}{%
    \begin{tabular}{lccccc}
        \toprule
        Method & C & B-4 & Id. & Evt. & Act. \\
        \midrule

        VideoLLM-Online~\cite{chen2024videollmonline} 
        & 15.962 & 11.962 & 0.000 & 10.870 & 22.283 \\

        MATR~\cite{song2024online} + Qwen3-VL~\cite{bai2025qwen3} 
        & 46.760 & 15.741 & 0.830 & 50.000 & 18.919 \\

        HAT~\cite{reza2024hat} + Qwen3-VL~\cite{bai2025qwen3} 
        & 50.900 & 16.490 & 0.794 & 50.000 & 22.581 \\

        MATR~\cite{song2024online} + MatchTime~\cite{rao2024matchtime} 
        & 73.880 & 16.896 & 4.979 & 50.000 & 18.378 \\

        StreamMind~\cite{ding2025streammind} 
        & 15.749 & 4.591 & 0.450 & 21.739 & 5.612 \\

        MATR~\cite{song2024online}+IAVC~\cite{xi2025player} 
        & 40.623 & 14.372 & 1.660 & 49.275 & 15.676 \\

        OnPoint~\cite{reza2026onpoint} + Qwen3-VL~\cite{bai2025qwen3} 
        & 31.300 & 11.272 & 0.459 & 32.609 & 15.962 \\

        \textbf{Ours}
        & \textbf{367.598}
        & \textbf{48.211}
        & \textbf{34.902}
        & \textbf{81.884}
        & \textbf{52.853} \\

        \bottomrule
    \end{tabular}%
    }
\end{table}

To eliminate the influence of different event matching sets and enable a
fair comparison of commentary generation quality, we further evaluate all
methods on the subset of events commonly matched by all methods.
Table~\ref{tab:common_events} compares all methods on this identical event
set. Our method achieves the best performance across all metrics,
outperforming the best-performing baseline on each metric by 293.718 CIDEr
points, 31.315 BLEU-4 points, and 29.923, 31.884, and 30.272 percentage
points in Identity, Event Type, and Action Accuracy, respectively.

\subsubsection{Qualitative Results}

Fig.~\ref{fig:qualitative_results} presents four representative
commentary examples. In the first three cases, our method closely matches
the ground truth, correctly recognizing the event classification, player identities,
and fine-grained actions, while the baselines produce clear errors in
event classification or player identity. The fourth example further illustrates
a more challenging case of a complex event chain. Although our method
successfully captures the overall structure of the putback\_chain, including
the missed shot, offensive rebound, and subsequent putback layup, it still
makes errors in the identities of the involved players. This indicates that,
despite strong performance on individual events, complex event chains such
as putback\_chain remain challenging because they require temporally
consistent association of multiple actions and player roles across
successive event stages.

\subsubsection{Human and LLM Evaluation}

We randomly sample 100 events commonly matched by all methods and evaluate
the generated commentaries using GPT-5.6 and two human raters. For human
evaluation, method identities are anonymized, and the outputs are scored
independently by each rater. All evaluations follow the same 0--5 rubric:
5 indicates that all core facts are correct; 4 allows one minor error; 3
indicates that the main event is correctly identified, but the commentary
contains one error that affects understanding; 2 captures only the broad
event with multiple errors; 1 indicates an incorrect main event; and 0
denotes an irrelevant output. Our method achieves an average GPT-5.6 score
of 3.24, compared with 2.11-2.45 for the baselines. The two human raters
assign average scores of 3.14 and 3.25 to our method, while the corresponding
baseline scores range from 1.94-2.10 and 1.74-2.21, respectively. Both
LLM-based and human evaluations demonstrate that our method generates
higher-quality basketball commentary.

\subsubsection{Ablation Study}

\begin{table}[!t]
    \centering
    \caption{Ablation results of the semantic modules in Stage II.
    C denotes CIDEr, while Id., Evt., and Act. denote identity,
    event-type, and fine-grained action accuracies (\%). Higher values
    indicate better performance for all metrics.}
    \label{tab:ablation}
    \scriptsize
    \renewcommand{\arraystretch}{1.10}
    \setlength{\tabcolsep}{3.0pt}
    \resizebox{\columnwidth}{!}{%
    \begin{tabular}{lccccc}
        \toprule
        Variant & C & B-4 & Id. & Evt. & Act. \\
        \midrule

        M0
        & 18.916 & 11.212 & 0.462 & 13.458 & 17.546 \\

        M0+M1
        & 71.456 & 26.703 & 0.146 & 69.896 & 39.979 \\

        M0+M1+M2
        & 231.513 & 38.334 & \textbf{28.647} & \textbf{72.008} & 40.723 \\

        M0+M1+M2+M3
        & \textbf{248.355}
        & \textbf{41.166}
        & 28.533
        & 71.387
        & \textbf{47.053} \\

        \bottomrule
    \end{tabular}%
    }
\end{table}

Table~\ref{tab:ablation} progressively adds M1--M3 to the Global Scene
Context Module (M0), with all variants evaluated on the same localized
events. The Ball Motion-Conditioned Event Perception Module (M1) improves
Event Accuracy from 13.458\% to 69.896\%, Action Accuracy from 17.546\% to
39.979\%, CIDEr from 18.916 to 71.456, and BLEU-4 from 11.212 to 26.703.
These results demonstrate that ball-centered motion and visual evidence
provide substantially stronger event semantics than global scene features
alone. Identity Accuracy slightly decreases from 0.462\% to 0.146\%,
which is reasonable because M1 focuses on event perception without
explicit player-identity supervision.

Adding the Event-Guided Player Identity Recognition Module (M2) sharply
increases Identity Accuracy from 0.146\% to 28.647\%, while CIDEr and
BLEU-4 increase from 71.456 to 231.513 and from 26.703 to 38.334,
respectively. CIDEr measures similarity to the reference commentary using
TF-IDF-weighted n-grams, and is therefore particularly sensitive to
informative and discriminative content such as player names. In contrast,
BLEU-4 measures modified n-gram precision up to 4-grams with a brevity
penalty, and thus benefits from correctly grounded identities that improve
exact lexical and local phrase matching with the reference commentary.

Finally, the Event-Guided Fine-Grained Action Recognition Module (M3)
improves Action Accuracy from 40.723\% to 47.053\%, while CIDEr further
increases from 231.513 to 248.355 and BLEU-4 from 38.334 to 41.166.
Fine-grained action recognition provides more discriminative action
semantics for CIDEr while producing more accurate action words and phrases
for BLEU-4. This improvement suggests that finer action distinctions
directly contribute to more precise and informative commentary generation.
Beyond these numerical changes, M3 more explicitly exposes fine-grained
action cues to the generator, making action semantics directly available
throughout the commentary generation process. This is particularly useful
for distinguishing events that share the same coarse event type but differ
in their specific action patterns. Identity Accuracy remains stable
(28.647\% to 28.533\%), whereas Event Accuracy changes slightly from
72.008\% to 71.387\%. Since M1 remains frozen, this small fluctuation mainly
reflects the influence of additional semantic prompts on the generated
commentary rather than a degradation of the underlying event perception.
Overall, the substantial gains in Identity and Action Accuracy, together
with the consistent improvements in CIDEr and BLEU-4, demonstrate that
accurate player grounding and fine-grained action understanding are key to
our commentary generation performance.
\section{Conclusions}

We introduce NBA\_Streaming, a benchmark for fine-grained basketball
commentary generation in continuous streams, together with a two-stage
framework combining completion-first localization and ball-centric
semantic grounding. Extensive experiments reveal the difficulty of
NBA\_Streaming, where existing baselines struggle with online timing,
factual grounding, and fine-grained description. Our framework consistently
improves over strong alternatives in event localization and commentary
generation, demonstrating the effectiveness of the proposed design.

Overall, NBA\_Streaming provides a challenging benchmark for realistic
streaming sports video understanding and commentary generation. Despite
the improvements achieved by our framework, challenges remain in modeling
complex event chains and consistently resolving player identities and
fine-grained actions across successive event stages. Future work will
explore more robust multimodal grounding and more efficient online
generation methods to further advance fine-grained commentary generation
in continuous video streams.

\bibliographystyle{IEEEtran}
\bibliography{reference}

\end{document}